\documentclass[journal]{IEEEtran}

\ifCLASSOPTIONcompsoc
  \usepackage[nocompress]{cite}
\else
  \usepackage{cite}
\fi

\usepackage{amsmath}
\usepackage{amsthm}
\usepackage{amssymb}
\usepackage{amsfonts}
\usepackage{enumitem}
\usepackage[pdftex]{graphicx}
\usepackage{bbm}
\usepackage{multirow}
\usepackage{xcolor}
\usepackage{hyperref}
\usepackage{pifont}
\usepackage{xr}
\usepackage{algorithm}
\usepackage[noend]{algpseudocode}
\usepackage{tablefootnote}
\algdef{SE}{Begin}{End}{\textbf{begin}}{\textbf{end}}

\DeclareMathOperator*{\argmin}{arg\,min}

\newcommand{\rch}[1]{\textcolor{black}{#1}}  
\newcommand{\och}[1]{\textcolor{black}{#1}}  

\begin{document}

\title{Training Deep Architectures Without End-to-End Backpropagation: \\ A Survey on the Provably Optimal Methods}

\author{Shiyu~Duan,
Jos\'{e} C. Pr\'{i}ncipe
\thanks{SD (\href{mailto:michaelshiyu3@gmail.com}{michaelshiyu3@gmail.com}) and JCP (\href{mailto:principe@cnel.ufl.edu}{principe@cnel.ufl.edu}) are with University of Florida, USA.}}

\markboth{}%
{Duan and Pr\'{i}ncipe: Training Deep Architectures Without End-to-End Backpropagation}

\maketitle

\begin{abstract}
  This tutorial paper surveys provably optimal alternatives to end-to-end backpropagation (E2EBP) --- the de facto standard for training deep architectures.
  \textit{Modular} training refers to strictly local training without both the forward and the backward pass, i.e., dividing a deep architecture into several nonoverlapping modules and training them separately without any end-to-end operation.
  Between the fully global E2EBP and the strictly local modular training, there are \textit{weakly modular} hybrids performing training without the backward pass only.
  These alternatives can match or surpass the performance of E2EBP on challenging datasets such as ImageNet, and are gaining increasing attention primarily because they offer practical advantages over E2EBP, which will be enumerated herein.
  In particular, they allow for greater modularity and transparency in deep learning workflows, aligning deep learning with the mainstream computer science engineering that heavily exploits modularization for scalability.
  Modular training has also revealed novel insights about learning and has further implications on other important research domains.
  Specifically, it induces natural and effective solutions to some important practical problems such as data efficiency and transferability estimation.
\end{abstract}

\begin{IEEEkeywords}
Deep learning, modular training, weakly modular training.
\end{IEEEkeywords}

\section{Introduction}
\IEEEPARstart{E}{nd}-to-end backpropagation (E2EBP) \cite{rumelhart1986learning} has been the de facto training standard for deep architectures since almost the inception of deep learning.
It optimizes the entire model simultaneously using backpropagated error gradients with respect to some target signal.
At each optimization step, E2EBP requires sending an input through the entire model to compute an error (an \textit{end-to-end forward pass}), and then taking gradient of the error with respect to all trainable parameters using the chain rule (an \textit{end-to-end backward pass}).

Historically, E2EBP was a major achievement that first enabled training multilayer neural networks effectively and, as a result, capitalizing on the expressiveness of universal function approximators \cite{cybenko1989approximation}. 
Many years of research has yielded E2EBP improvements that enabled training deep neural networks with thousands of layers and millions of units. 
These include advances in both hardware and implementation such as large-scale distributed training with massive datasets, residual connections inside networks to facilitate end-to-end gradient flow, and so on. 
And E2EBP has been intertwined with the success of deep learning \cite{krizhevsky2012imagenet,devlin2019bert,silver2016mastering}.
The versatility, practicality, and theoretical optimality (in terms of first-order gradient) of E2EBP made it the de facto standard. 

However, E2EBP is not perfect and has shortcomings that create practical difficulties, especially when scaled to large architectures.
First, gradients may vanish or explode when propagated through too many layers, and much care is needed to avoid such suboptimal training dynamics \cite{he2016deep,zhang2019gradient}.
As another drawback, end-to-end training requires optimizing the model as a whole, which does not allow for modularized workflows and turns models into ``black boxes'', making it difficult for the users to debug models or extract any useful insight on the learning task from trained models \cite{duan2021modularizing}.
This inflexibility also means that training larger models with E2EBP would necessitate more computational resources, raising the bar to adopt deep learning technologies and creating concerns over their energy consumptions and environmental impacts \cite{gomez2020interlocking,wang2021revisiting}.
Furthermore, E2EBP can scale poorly to deep models due to less-than-ideal loss landscapes and spurious correlations \cite{ioffe2015batch}.
The convergence may become slow and the reached local minimum may be suboptimal.

More subtle are the issues of how effective and how much information is preserved in the backpropagation of error gradient. 
From the optimization standpoint, as long as the model is differentiable, a gradient can be defined and an extremum (or saddle point) of the loss function can be reached by simply following the gradient, which is what E2EBP amounts to.
However, in many real-world scenarios, the user may need more control on the characteristics of the internal representations to improve robustness, generalization, etc.
In E2EBP, the error can only be computed after sending the internal representations through the usually many downstream layers and eventually into the output space.
As a result, the effect of the output loss function penalty on a hidden layer cannot be effectively controlled because of the nonlinear layers in between \cite{lee2015deeply}.

The above practical difficulties of E2EBP can make realizing the full potential of deep learning difficult in applications.
Therefore, it is rather important to seek alternatives to E2EBP that can preserve the good properties of E2EBP while improving its shortcomings.
Modular and weakly modular training are such alternatives.
\rch{This paper} uses the term \textit{modular training} to refer to training schemes that do not require end-to-end forward or backward pass, and \textit{weakly modular training} for training schemes that need end-to-end forward pass but not end-to-end backward pass.

In simple terms the possibilities to avoid E2EBP are all related to the use of the target signal in training.
E2EBP creates the error at the output and backpropagates it to each layer in the form of gradients.
Instead, one can use the target at each layer to train it locally with a proxy objective function, in a modular way without backpropagating the error gradients end-to-end. 
Alternatively, one can approximate the inverse targets or local gradients to yield weak modularity. 

Besides solving the above practical issues of E2EBP, modular training offers novel insights and practical implications to other research domains such as transferability estimation \cite{duan2021modularizing} and optimization.
The main limitation of modular and weakly modular training is that not all such methods can provide similar theoretical optimality guarantees in generic settings as E2EBP does.

There is a large body of work studying modular or weakly modular training on modern deep neural networks  \cite{duan2019kernel,duan2021modularizing,lee2015difference,nokland2019training,jaderberg2017decoupled,czarnecki2017understanding,lansdell2020learning,bengio2014auto,belilovsky2019greedy,pogodin2020kernelized,meulemans2020theoretical,belilovsky2020decoupled,manchev2020target,nokland2016direct,lillicrap2016random,moskovitz2018feedback,samadi2017deep,krotov2019unsupervised,qin2021contrastive,mostafa2018deep,marquez2018deep,xiao2018biologically,carreira2014distributed,balduzzi2014kickback,ororbia2020continual,guerguiev2019spike,kunin2020two,launay2019principled,taylor2016training,podlaski2020biological,ororbia2019biologically,laborieux2020scaling,scellier2017equilibrium,veness2019gated,ma2020hsic,huang2018learning,obeid2019structured,song2020can,whittington2017approximation,pehlevan2018similarity,malach2018provably,gu2020fenchel,zhang2017convergent,askari2018lifted,lau2018proximal,carreira2016parmac,li2019lifted,zhang2016efficient,zeng2019global,marra2020local,raghavan2020distributed,bartunov2018assessing,choromanska2019beyond,lukasiewicz2020can,baldi2018learning,liao2016important,bengio2020deriving,lowe2019putting,wang2021revisiting}.
However, only a few of the existing schemes have been shown to produce competitive performance on meaningful benchmark datasets.
Even fewer provided some form of optimality guarantee under reasonably general settings.
And despite the surging interest, no recent existing work provides a survey on these provably optimal modular and weakly modular training methods.

The goal of this paper is exactly to review existing work on avenues to train deep neural networks without E2EBP, so as to push the field to new highs.
\rch{This paper mainly focuses on provably optimal methods, reviewing them extensively.
Other popular families of methods are also discussed after.}
\rch{Formally, this paper defines \textit{provably optimal} training methods to be those that can be proven to yield optimal solutions to the given training objective in reasonably general settings, with some level of flexibility in how ``optimal'' is interpreted (for example, optimal with respect to first-order gradient and optimal with respect to first-order and second-order gradients are both valid interpretations because both are relevant in practice).}

Why \rch{put more emphasis on} the provably optimal approaches?
For practitioners, to the best of our knowledge, only provably optimal methods have been shown to produce performance comparable to E2EBP on meaningful benchmark datasets \cite{belilovsky2019greedy,belilovsky2020decoupled,nokland2019training,duan2021modularizing,wang2021revisiting}, whereas others have been greatly outperformed by E2EBP \cite{bartunov2018assessing}.
For theoreticians, provably optimal methods offer theoretical insights \rch{that} the others do not. 
In particular, from a learning theoretic perspective, the fundamental job of a learning method is to effectively find a solution that best minimizes a given objective \cite{shalev2014understanding}.
E2EBP is obviously capable of doing this. 
And if a non-E2E alternative is not guaranteed to find a minimizer to the training objective, it is not a very compelling alternative to E2EBP because it cannot get the most basic job done.
In summary, our choice of reviewing only the probably optimal methods is motivated both by empirical performance and theoretical value.
And to dive deep into the theory, we had to somewhat sacrifice the breadth of this survey and focus \rch{mainly} on these methods.

These provably optimal methods can be categorized into three distinct abstract algorithms: Proxy Objective (modular), Target Propagation (weakly modular), and Synthetic Gradients (weakly modular).
And for each abstract algorithm, its popular instantiations, optimality guarantees, advantages and limitations, and potential implications on other research areas are discussed.
Some future research directions will also be sketched out.
A summary of all methods is provided in Table~\ref{table1}, their best reported results on standard benchmark datasets are listed in Table~\ref{table2}, and illustrations are given in Fig.~\ref{fig1}.

\rch{The other families of methods this paper reviews include Feedback Alignment (weakly modular) and Auxiliary Variables (weakly modular).
Due to a lack of established optimality guarantee or other limitations, we review these methods somewhat less extensively after the provably optimal ones.
Reported results from popular instantiations of these methods on standard benchmark datasets are given in Table~\ref{table2}.
}

This tutorial paper is intended to serve as an entry point to readers that either wish to engage in research in modular or weakly modular learning or simply would like to use these training methods in their applications.
The readers should be able to gain a holistic view on the existing methods as well as a full understanding on how each of them work.
However, detailed discussions and optimality proofs should be found in the cited original papers.
This paper assumes basic knowledge on the core concepts in deep learning such as knowing what a deep neural network is and how E2EBP works.

In Section~\ref{sec2}, \rch{the settings considered by this survey are described}.
The formal presentation \rch{on provably optimal methods} is in Section~\ref{sec3}.
Specifically, Proxy Objective methods will be discussed in Section~\ref{sec3:po}, Target Propagation in Section~\ref{sec3:tp}, and Synthetic Gradients in~\ref{sec3:sg}.
\rch{Feedback Alignment and Auxiliary Variables are reviewed in Section~\ref{other_methods}.}

\section{The Settings}
\label{sec2}

\rch{This survey} considers the task of classification using feedforward neural networks, which is a set-up in which all reviewed methods have been shown to work in the respective papers.
Some of the methods have been evaluated on other architectures such as recurrent networks and for other tasks such as regression.
Some other methods can be extended to more models and tasks, but there are no papers known to us that have provided empirical evidence on the practical validity of such extensions.
In the rest of this paper, the term \textit{module} is used to mean a composition of an arbitrary number of network layers.



Consider a two-module network \(f\left(\cdot, \theta_1, \theta_2\right) = f_2\left(f_1\left(\cdot, \theta_1\right), \theta_2\right)\) for simplicity, where \(\theta_1\) represents the trainable parameters of the input module \(f_1\) and \(\theta_2\) the output module \(f_2\).
Each method presented can be trivially extended to training with more than two modules by analyzing one pair of modules at a time.

Note that \(f_i\) is not only defined by its trainable parameters, but also the non-trainable ones.
For example, a fully-connected layer on \(\mathbb{R}^d\) with \(p\) nodes and some nonlinearity \(\sigma:\mathbb{R}^p\to\mathbb{R}^p\) can be written as \(f_i\left(\cdot, W_i\right):\mathbb{R}^d\to\mathbb{R}^p:x\mapsto \sigma\left(W_i x\right)\), where \(W_i\) is its trainable weight matrix.
In this case, \(\theta_i\) is \(W_i\), and that this layer is fully-connected together with the nonlinearity \(\sigma\) constitutes the non-trainable parameters.
For \(f_i\), its non-trainable parameters are denoted as \(\omega_i\).
\rch{The dependence of \(f_i\) on the non-trainable parameters \(\omega_i\) will not be explicitly written out as this survey only focuses on the training of the model.}

Let the data be \((X, Y)\)\footnote{\rch{For data, random elements are denoted using capital letters and lower-case letters are reserved for their observations (actual data collected).}}, with \(X\) being the input example --- a random element on \(\mathbb{X}\subset\mathbb{R}^d\) for some \(d\), and \(Y\) its label --- a random variable on \(\mathbb{Y}\subset\mathbb{R}\).
Suppose a loss function \(\ell:\mathbb{Y}\times\mathbb{Y}\to\mathbb{R}\) is given.
Define the risk as \(R\left(f\left(\cdot, \theta_1, \theta_2\right), X, Y\right) = E_{(X, Y)}\ell\left(f\left(X, \theta_1, \theta_2\right), Y\right)\)
The goal is to find some \(\theta_1^\prime, \theta_2^\prime\) such that \(\left(\theta_1^\prime, \theta_2^\prime\right)\in\argmin_{\left(\theta_1, \theta_2\right)} R\left(f\left(\cdot, \theta_1, \theta_2\right), X, Y\right)\).
\rch{Suppose a training set \(S = \left\{x_i, y_i\right\}_{i=1}^n\) is given.}
And in practice, the risk can be estimated by an objective function, e.g., the sample average of the loss evaluated on \(S\) or the sample average together with a regularization term.
Let an objective function be \(L\left(f, \theta_1, \theta_2, S\right)\), and the goal in practice is to minimize this objective in \(\left(\theta_1, \theta_2\right)\).

\textit{End-to-end forward pass} refers to sending input through the entire model, i.e., evaluating \(f\left(x, \theta_1, \theta_2\right)\) for some input \(x\).
And \textit{end-to-end backward pass} means taking gradient of the training objective with respect to all trainable parameters, i.e., evaluating the partial derivatives of \(L\) with respect to both \(\theta_1\) and \(\theta_2\) using the chain rule.
\rch{This survey} uses the term \textit{modular training} to refer to training schemes that do not require end-to-end forward pass or backward pass, and \textit{weakly modular training} for training schemes that need end-to-end forward pass but not end-to-end backward pass.

\section{Provably Optimal Learning Schemes}
\label{sec3}

In this section, the reviewed modular and weakly modular training methods are presented.
Modular methods will be discussed first in Section \ref{sec3:modular}.
Weakly modular methods will be presented in Section~\ref{sec3:ne2e}.

\begin{figure*}[t]
    \centering
    \includegraphics[width=\textwidth]{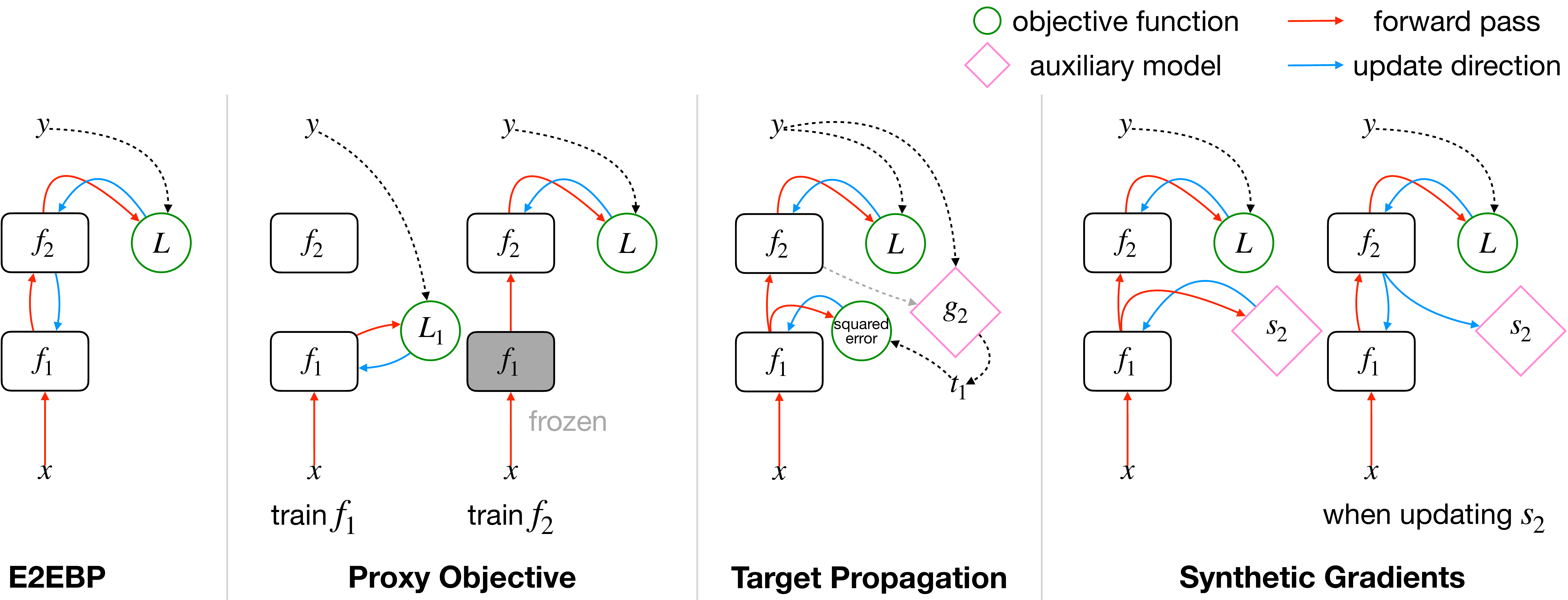}
    \caption{
      E2EBP, modular, and weakly modular training schemes in the case where a model is trained as two modules.
      Proxy Objective methods leverage a strictly local proxy objective \(L_1\) for training the input module.
      Target Propagation approximates an ``inverse'' of the output module \(f_2\) using an auxiliary learnable inverse model \(g_2\), then backpropagates a target \(t_1\) instead of gradient (the connection from \(f_2\) to \(g_2\) is optional).
      Synthetic Gradients methods approximate local gradient with an auxiliary trainable gradient model \(s_2\).
    }
    \label{fig1}
\end{figure*}

\begin{table*}[t]
\caption{Provably optimal modular and weakly modular training methods.}
\label{table1}
\centering
\begin{tabular}{|c|c|c|c|c|}
\hline
\textbf{Training Scheme} & \textbf{Main Idea} & \textbf{Auxiliary Model} & \textbf{Forward Pass} & \textbf{Backward Pass}\\
\hline
Proxy Objective & Use a local proxy objective function & Not necessary & Not required & Not required \\
\hline
Target Propagation & Approximate ``inverse'' of downstream layers, backpropagate targets & Required & Required & Not required \\
\hline
Synthetic Gradients & Approximate local gradients & Required & Required & Required \\
\hline
\end{tabular}
\end{table*}

\begin{table*}[t]
\caption{Performant instantiations of reviewed methods and best reported test accuracy on standard benchmarking datasets. Note that each result entry should only be compared against the corresponding E2EBP baseline due to the potentially different test settings across entries. VGG-11B and VGG-8B are customized VGG-11 and VGG-8, respectively. Only Proxy Objective methods can match the performance of E2EBP on competitive networks.}
\label{table2}
\centering
\begin{tabular}{|c|c|c|c|c|c|c|}
\hline
\textbf{Training Scheme} & \textbf{Instantiation} & \textbf{Test Dataset} & \textbf{Network} & \textbf{Modularity} & \textbf{Acc. (\%)} & \textbf{E2EBP Acc. (\%)}\\
\hline
\multirow{10}{*}{Proxy Objective} & \cite{belilovsky2019greedy} & ImageNet & VGG-11 & Trained as 3-layer modules & 67.6 (Top-1)/88.0 (Top-5) & 67.9/88.0 \\
\cline{2-7}
& \multirow{3}{*}{\cite{belilovsky2020decoupled} (synchronous)} & \multirow{3}{*}{ImageNet} & VGG-13 & Trained as 4 modules & 67.8/88.0 & 66.6/87.5 \\
& & & VGG-19 & Trained as 2 modules & 70.8/90.2 & 69.7/89.7 \\
& & & ResNet-152 & Trained as 2 modules & 74.5/92.0 & 74.4/92.1 \\
\cline{2-7}
& \multirow{4}{*}{\cite{nokland2019training}} & CIFAR-10 & VGG-11B & \multirow{4}{*}{Trained as 1-layer modules} & 96.03 & 94.98 \\
& & CIFAR-100 & VGG-11B & & 79.9 & 76.3 \\
& & SVHN & VGG-8B & & 98.26 & 97.71 \\
& & STL-10 & VGG-8B & & 79.49 & 66.92 \\
\cline{2-7}
& \multirow{2}{*}{\cite{duan2021modularizing}} & \multirow{2}{*}{CIFAR-10} & ResNet-18 & \multirow{2}{*}{Trained as 2 modules} & 94.93 & 94.91 \\
& & & ResNet-152 & & 95.73 & 95.87 \\
\cline{2-7}
& \multirow{4}{*}{\cite{wang2021revisiting}} & CIFAR-10 & \tiny{DenseNet-BC-100-12} & \multirow{4}{*}{Trained as 2 modules} & 95.26 & 95.39 \\
& & SVHN & ResNet-110 & & 96.85 & 96.93 \\
& & STL-10 & ResNet-110 & & 79.01 & 77.73 \\
& & ImageNet & ResNeXt-101 & & 79.65/94.72 & 79.36/94.60 \\
\hline
\multirow{2}{*}{Target Propagation} & \multirow{2}{*}{\cite{meulemans2020theoretical}} & \multirow{2}{*}{CIFAR-10} & Small MLP & \multirow{2}{*}{Trained layerwise} & 49.64 & 54.40 \\
& & & Small CNN & & 76.01 & 75.62 \\
\hline
\multirow{2}{*}{Synthetic Gradients} & \multirow{2}{*}{\cite{lansdell2020learning}} & CIFAR-10 & \multirow{2}{*}{Small CNN} & \multirow{2}{*}{Trained layerwise} & 74.8 & 76.9 \\
& & CIFAR-100 & & & 48.1 & 51.2 \\
\hline
\multirow{1}{*}{\rch{Feedback Alignment}} & \multirow{1}{*}{\cite{xiao2018biologically}} & \rch{ImageNet} & \multirow{1}{*}{\rch{AlexNet}} & \multirow{1}{*}{\rch{Trained layerwise}} & \rch{47.57/23.68} & \rch{49.15/25.01} \\
\hline
\multirow{2}{*}{\rch{Auxiliary Variables}} & \multirow{2}{*}{\cite{zeng2018block}} & \rch{MNIST} & \multirow{2}{*}{\rch{Small MLP}} & \multirow{2}{*}{\rch{Trained layerwise}} & \rch{95.68} & \rch{95.33} \\
& & \rch{CIFAR-10} & & & \rch{44.96} & \rch{46.99} \\
\hline
\end{tabular}
\end{table*}

\subsection{Modular Learning Schemes}
\label{sec3:modular}

\subsubsection{Proxy Objective}
\label{sec3:po}

This strategy amounts to finding a \textit{proxy objective function} \(L_1\left(f_1, \theta_1, \omega_2, S\right)\) that can be used to train the input module.
The important thing to note is that this proxy objective does not involve the trainable parameters \(\theta_2\), which enables decoupling the training of \(f_1\) and \(f_2\) completely.
Methods based on a proxy objective can be abstracted into Algorithm~\ref{alg:proxy} (also see Fig.~\ref{fig1} for an illustration).
Note that these methods focus on finding \(L_1\), but not the actual optimization on \(L_1\) or \(L\) --- any off-the-shelf optimizer such as stochastic gradient descent can be used here.

\begin{algorithm}[t]
  \caption{Proxy Objective}
  \label{alg:proxy}
  \begin{algorithmic}[1]
    \Require A proxy objective function \(L_1\)
    \Begin
    \State Find \(\theta_1^\star\in\argmin_{\theta_1}L_1\left(f_1, \theta_1, \omega_2, S\right)\)
    \State Find \(\theta_2^\star\in\argmin_{\theta_2}L\left(f, \theta_1^\star, \theta_2, S\right)\)
    \State \Return \(\left(\theta_1^\star, \theta_2^\star\right)\)
    \End
  \end{algorithmic}
\end{algorithm}

\paragraph{Instantiations}
Different variations of Proxy Objective methods differ mainly in how they choose the proxy objective \(L_1\).
Essentially, the existing proxy objectives can all be interpreted as characterizations of hidden representation separability.
Then training on \(L_1\) encourages \(f_1\) to improve separability of its output representations, making the classification problem simpler for \(f_2\).
There are mainly two schemes for quantifying this separability.
\begin{itemize}
  \item \textbf{Separability through feature space distance/similarity} 
  Separability can be quantified via distance/similarity of learned hidden representation vectors.
  In \cite{duan2019kernel,duan2021modularizing,wang2021revisiting}, this idea has been explored.
  \cite{duan2019kernel,duan2021modularizing} proposed methods that do not require training auxiliary models, whereas in \cite{wang2021revisiting}, auxiliary networks were needed for transforming representations.

  \cite{duan2021modularizing} proposed to quantify output data separability for \(f_1\) in terms of distances between its output data representation pairs in their native feature space.
  Then \(L_1\) should encourage \(f_1\) to map example pairs from distinct classes further apart for better separability.
  Specifically, suppose \(f_1\) ends with a nonlinearity \(\phi\) mapping into some inner product space such as ReLU, that is, \(f_1\left(\cdot, \theta_1\right) = \phi\circ g_1\left(\cdot, \theta_1\right)\), where \(g_1\) is some other function, any loss function that can help learn \(\theta_1\) such that \(g_1\) maximizes 
  \begin{equation}
    \label{eq1}
    \left\|\phi\left(g_1\left(x_i, \theta_1\right)\right) - \phi\left(g_1\left(x_j, \theta_1\right)\right)\right\|
  \end{equation}  
  for all pairs of \(x_i, x_j\) from \(S\) with \(y_i\neq y_j\) can be used as a proxy objective, where the norm is the canonical norm induced by the inner product.
  More generally, this distance can be substituted by other distance metrics or similarity measures.

  In addition to encouraging \(f_1\) to separate examples from distinct classes, one can additionally modify \(L_1\) such that \(f_1\) is encouraged to map examples from identical classes closer, making the pattern more separable for \(f_2\).
  To implement this idea, \cite{duan2019kernel} suggested minimizing 
  \begin{equation}
    \label{eq2}
    \left\|\phi\left(g_1\left(x_i, \theta_1\right)\right) - \phi\left(g_1\left(x_j, \theta_1\right)\right)\right\|
  \end{equation}  
  for all pairs of \(x_i, x_j\) from \(S\) with \(y_i = y_j\) in addition to the primary proxy (Eq. ~\ref{eq1}).

  In \cite{wang2021revisiting}, the authors proposed to attach an auxiliary representation-learning convolutional neural network (CNN) \(h_1\left(\cdot, \alpha_1\right)\) (\(\alpha_1\) denotes its trainable parameters) to \(f_1\) before computing feature distance/similarity.
  This auxiliary model mapped the hidden representation to another representation space, and the proxy objective can be any objective function that drives \(h_1\left(f_1\left(x_i, \theta_1\right), \alpha_1\right)\) and \(h_1\left(f_1\left(x_j, \theta_1\right), \alpha_1\right)\) more (less) similar if \(y_i = (\neq) y_j\).
  \(\alpha_1, \theta_1\) were trained jointly to optimize the proxy objective, and \(h_1\) was discarded during inference time.
  Compared to \cite{duan2019kernel,duan2021modularizing}, the addition of an auxiliary model gives user the flexibility to choose the dimensionality of the feature space in which the proxy objective is computed regardless of the given network layer dimension.
  The downside is that this auxiliary model requires extra resources to design and train.

  Note that the training method in \cite{duan2019kernel} was initially established for the so-called ``kernel networks''.
  But \cite{duan2021modularizing} later showed that neural networks are special cases of kernel networks, therefore making all results in \cite{duan2019kernel} applicable to neural networks as well.

  \item \textbf{Separability through auxiliary classifier performance} Instead of expressing data separability via distance in feature space as above, \cite{belilovsky2019greedy,wang2021revisiting} attached an auxiliary CNN classifier to the output of \(f_1\) and used its accuracy as an indicator of data separability.
  To train \(f_1\) to improve data separability, the authors proposed to use a classification loss such as cross-entropy on this auxiliary classifier and \(f_1\) as a proxy objective for training \(f_1\).
  This auxiliary classifier was trained together with \(f_1\) to minimize the proxy objective and was discarded at test time.

  \cite{mostafa2018deep,marquez2018deep} explored a similar idea but used different auxiliary classifiers.
  Specifically, the auxiliary classifiers in \cite{mostafa2018deep} were single-layer fully-connected networks with random, fixed weights, and in \cite{marquez2018deep}, the auxiliary classifiers were two-layer, fully-connected, and trainable.
  Using fixed auxiliary classifiers produces less performant main classifiers, as demonstrated in \cite{mostafa2018deep}.

  In \cite{wang2021revisiting}, expressing data separability through similarity in feature space almost always outperformed doing so through performance of auxiliary classifiers in terms of final network performance.

  \item As a combination of the earlier two approaches, one may quantify data separability in terms of a combination of both feature space distance and auxiliary classifier accuracy.
  In \cite{nokland2019training}, two auxiliary networks were used to facilitate the training of \(f_1\).
  The first network, denoted \(h_1\left(\cdot, \alpha_1\right)\), was a convolutional representation-learning module (\(\alpha_1\) denotes its trainable parameters) and the second was a linear classifier denoted \(q_1\left(\cdot, \beta_1\right)\) (\(\beta_1\) denotes its trainable parameters).
  The proxy objective was a weighted combination of a ``similarity matching loss'' and a cross-entropy loss. 
  The similarity matching loss was the Frobenius norm between the cosine similarity matrix of \(h_1\left(f_1\left(\cdot, \theta_1\right), \alpha_1\right)\) evaluated on a batch of training examples and the cosine similarity matrix of their one-hot encoded labels.
  The cross-entropy loss was on \(q_1\left(f_1\left(\cdot, \theta_1\right), \beta_1\right)\).
  \(f_1, h_1, q_1\) were trained jointly to minimize the proxy objective, and \(h_1, q_1\) were discarded at test time.

  Minimizing the similarity matching loss can be interpreted as maximizing distances between pairs of examples from distinct classes and minimizing those between pairs from identical classes.
  Therefore, this loss term can be interpreted as quantifying data separability in terms of feature space distance.
  The cross-entropy loss and the use of an auxiliary classifier can be viewed as expressing data separability using auxiliary classifier accuracy. 
  Thus, this method can be seen as a combination of the earlier two approaches.

  \item In an effort to further reduce computational complexity and memory footprint for the greedy training method in \cite{belilovsky2019greedy}, \cite{belilovsky2020decoupled} proposed two variants. 
  In the synchronous variant, for each training step, an end-to-end forward pass was performed, then all modules were trained simultaneously, each minimizing its own proxy objective as proposed in \cite{belilovsky2019greedy}.
  The algorithmic stability for this variant was proven both theoretically and empirically.
  In the other asynchronous variant, no end-to-end forward pass was needed.
  Instead, all modules were trained simultaneously in an asynchronous fashion.
  This was made possible by maintaining a replay buffer for each module that contained stashed past outputs from the immediate upstream module.
  Then each module received its input from this replay buffer instead of actual current output from its upstream module, eliminating the need for end-to-end forward pass.
  Each module was still trained to minimize its own proxy objective in this asynchronous variant.
  The synchronous variant is not fully modular, since end-to-end forward pass is needed.

  The advantage of these variants is that, especially with the asynchronous variant, one can implement highly parallelized pipelines to train modules simultaneously instead of sequentially, further improving training speed and memory efficiency.
  Simultaneous training of all modules cannot be done with other Proxy Objective methods since the training of \(f_2\) requires outputs from a trained \(f_1\).

  \item One practical issue observed in the above modular training works is that the performance usually drops when a given network is split into too many thin modules \cite{wang2021revisiting}.
  \cite{wang2021revisiting} hypothesized that this was caused by greedy training ``collapsing'' task-relevant information too aggressively in the early modules.
  And the authors proposed a new (intractable \rch{due to the need for estimating a maximizer of a mutual information term}) proxy objective motivated from an information theoretic perspective.
  In practice, their proposal amounted to an add-on term to the previously discussed proxy objectives.
  This add-on term was a reconstruction loss of a CNN decoder head attempting to reconstruct \(x\) using \(f_1(x, \theta_1)\) as input.
  The decoder and \(f_1\) were jointly trained to minimize the proxy objective and this extra reconstruction loss term.
  Conceptually, it is easy to see that this extra reconstruction loss encourages \(f_1\) to retain information about the input \(x\), countering how the earlier proxy objectives drive \(f_1\) to keep information only about the label \(y\).
  According to the authors, information about input may not be immediately helpful for improving separability in the hidden layers (as opposed to information about label), but it may help overall network performance. 

\end{itemize}


\paragraph{Optimality Guarantees}

\cite{duan2021modularizing} gave an in-depth analysis on the optimality of training using the proxy objective proposed therein (Eq. \ref{eq1}).
In a modular training setting, the optimal set of input module parameters are the ones for which there exists output module(s) such that they combine into a minimizer for the overall objective.
Mathematically, this optimal set is given as
\begin{equation}
  \Theta_1^\star:= \left\{\theta_1\Bigg\vert \exists\theta_2\text{ s.t. }\left(\theta_1, \theta_2\right)\in\argmin_{\left(\theta_1, \theta_2\right)} L\left(f, \theta_1, \theta_2, S\right)\right\}.
\end{equation}
It was shown, under certain assumptions on \(L\), \(\phi\), and \(f_2\), that \(\argmin_{\theta_1}L_1\left(f_1, \theta_1, \omega_2, S\right)\subset\Theta_1^\star\) (Theorem 4.1 in \cite{duan2021modularizing}), justifying the optimality of training \(f_1\) to minimize the proxy objective.
The assumptions on \(L\) and \(\phi\) are mild.
In particular, the result works with popular classification objectives such as cross-entropy.
However, the assumption on \(f_2\) requires that it is a linear layer.
Overall, this optimality guarantee covers the case where a network is trained as two modules, with the output linear layer as the output module and everything else as the input module.
Optimality guarantee for more general training settings (training as more than two modules) was provided in \cite{duan2019kernel} (Theorem 2), but under stronger assumptions on \(\phi\) (see discussion immediately preceding the theorem).

One thing to note with this analysis is that the proxy objective is only picked to have its minima aligned with those of the overall objective. 
For smooth loss landscapes, solutions in a small neighborhood around a minimum should still effectively make the overall objective function value small.
But in general, the behavior of the proxy objective away from its minima is not constrained to align with that of the overall objective.
This implies that the solutions learned with this method may be suboptimal when the module being trained does not have enough capacity to minimize its proxy objective to a reasonable degree.
In practice, \cite{nokland2019training,wang2021revisiting} showed that effective training with low-capacity modules can be achieved by essentially enhancing the method in \cite{duan2021modularizing} with the idea of \cite{belilovsky2019greedy} or an add-on reconstruction term.
A theoretical understanding of the instantiation proposed in \cite{nokland2019training} is still lacking.

In \cite{wang2021revisiting}, the authors motivated their proxy objectives from an information theoretic perspective.
Specifically, they showed that a proxy objective formed by combining their reconstruction loss term with either feature space similarity/distance or auxiliary classifier performance can be interpreted as simultaneously maximizing (1) mutual information between hidden representation and input and (2) mutual information between hidden representation and label. 
However, the authors did not establish a connection between this mutual information maximization scheme and minimizing the overall training objective for the network.
Nevertheless, the strong empirical performance shown in the paper hints that such a connection should exist.

\cite{belilovsky2019greedy} provided a brief analysis on the optimality of their method and sketched out ideas that can potentially lead to optimality results.
But no concrete optimality guarantee was established.
\cite{malach2018provably} analyzed in detail a more complicated variant of the method in \cite{belilovsky2019greedy}.
It was shown that training algorithm was able to produce a correct classifier under a very strong data generation assumption.
However, the proposed variant in \cite{malach2018provably} also imposed architectural constraints on the model and was not accompanied by empirical evidence that it would work well with competitive network architectures.

\och{Although not originally established for modular training, some results from similarity learning can be viewed as optimality guarantees for Proxy Objective methods.
\((\epsilon, \gamma, \tau)\)-good similarity characterizes ideal hidden representations for which a downstream linear separator achieving small classification risk exists\cite{balcan2008improved,bellet2012similarity}.
Proxy objectives can be so designed that they drive the hidden layers towards \((\epsilon, \gamma, \tau)\)-good representations.}

\paragraph{Advantages}

\begin{itemize}
  \item Proxy Objective methods have by far the best empirical performance among alternatives to E2EBP (see Table~\ref{table2} and \cite{bartunov2018assessing}).
  To the best of our knowledge, they are the only family of modular or weakly modular training methods that have been shown to produce similar or even better performance compared to E2EBP on challenging benchmark datasets such as CIFAR-10 \cite{krizhevsky2009learning,duan2021modularizing,nokland2019training,belilovsky2020decoupled,wang2021revisiting}, CIFAR-100, SVHN \cite{netzer2011reading,wang2021revisiting}, STL-10 \cite{coates2011analysis,nokland2019training,wang2021revisiting}, and ImageNet \cite{russakovsky2015imagenet,deng2009imagenet,belilovsky2019greedy,belilovsky2020decoupled,wang2021revisiting}.
  \item Proxy Objective methods are arguably the simplest to use in practice compared to the other methods reviewed.
  In particular, unlike the other methods, Proxy Objective methods do not always require learning auxiliary models \cite{duan2019kernel,duan2021modularizing}.
  \item Proxy Objective methods give the user full and direct control over the hidden representations.
  When training a hidden module with a proxy objective, the supervision and potentially any side information that the user wishes to inject into the module can be directly passed to it without being propagated through the downstream module first.
\end{itemize}

\paragraph{Current Limitations and Future Work}  

\begin{itemize}
  \item Despite the strong empirical performance of Proxy Objective methods and the existence of some optimality proofs, optimality analysis for the various instantiations is far from complete.

  It is unclear how the optimality analysis performed for feedforward models in classification \cite{duan2019kernel,duan2021modularizing} can be extended to non-feedforward architectures, other objective functions in classification, or regression tasks.
  
  Optimality guarantees for training settings beyond the simple two-module one in, e.g., \cite{duan2021modularizing}, are lacking.
  Although, \cite{nokland2019training,wang2021revisiting} provided strong empirical evidence that modular training with few architectural assumptions and arbitrarily fine network partitions can still produce E2EBP-matching performance.
  Therefore, a theoretical analysis on the method in \cite{nokland2019training} (essentially combining \cite{duan2021modularizing} and \cite{belilovsky2019greedy}) and \cite{wang2021revisiting} (adding an extra reconstruction term) may be of interest.
  \item While much work has been done on studying the hidden representations learned by E2EBP, little is known about those learned by modular training.

  There are at least two directions in which this topic can be pursued. 
  First, deep learning models have been known to create ``hierarchical'' internal representations under E2EBP --- a feature widely considered to be one of the factors contributing to their success \cite{bengio2013representation}.
  As deep architectures trained with Proxy Objective methods have been shown to provide state-of-the-art performance on challenging datasets such as ImageNet \cite{belilovsky2020decoupled,wang2021revisiting}, it would therefore be interesting to dissect these models and study the characteristics of the internal representations learned.
  Notably, \cite{wang2021revisiting} already pointed out that some existing Proxy Objective methods do not learn representations that are ``hierarchical'' enough, leading to poor performance when a given model is trained as too many thin modules.

  As another direction, while there has been work studying using proxy objectives as side objectives along with E2EBP to enhance generalization of the model \cite{lee2015deeply}, the effects of using proxy objectives alone to inject prior knowledge into the model are understudied in the context of modular training.
  It is possible that such practice can yield stronger effects due to its more direct nature.
  And future work may study its impact on model generalization, adversarial robustness \cite{goodfellow2014explaining}, etc.
\end{itemize}

\paragraph{Further Implications on Other Research Domains}
\label{others}

Proxy Objective methods have profound implications on deep learning.

\begin{itemize}
  \item \textbf{Modularized workflows:}
  Fully modular training unlocks modularized workflows for deep learning.
  As argued in \cite{duan2021modularizing,gomez2020interlocking,jaderberg2017decoupled}, such workflows can significantly simplify the usually arduous procedure of implementing performant deep learning pipelines by allowing effective divide-and-conquer.
  Indeed, most engineering disciplines such as software engineering consider modularization as an integral part of any workflow for enhanced scalability.
  And modular training brings the freedom to embrace full modularization to deep learning engineering.
  \item \textbf{Transfer learning:}
  As an example on how modularized workflows can be advantageous in practice, \cite{duan2021modularizing} showed that a proxy objective can be naturally used as an estimator for module reusability estimation in transfer learning.
  Since a module that better minimizes a proxy objective can be used to build a network that better minimizes the overall objective, evaluating a proxy objective value on a batch of target task data can be used as an indicator on the transfer performance of a trained network body. 
  This idea can be further extended to provide a model-based solution to the task transferability estimation problem --- the theoretical problem behind reusability estimation.
  \cite{duan2021modularizing} provided some empirical evidence on the validity of this simple approach, but rigorous comparisons against other existing (and often more complicated) methods from transfer learning literature remains a future work.
  \item \textbf{Label requirements:}  
  The Proxy Objective instantiations proposed in \cite{duan2021modularizing,wang2021revisiting} use only labeled data of the form \(\left(x_i, x_j, \mathbbm{1}_{\{y_i = y_j\}}\right)\) for training the hidden layers, where \(\mathbbm{1}\) denotes the indicator function.
  And according to the optimality result in \cite{duan2021modularizing}, labeled data of this form is sufficient for training the hidden layers --- the specific class of \(x_i\) or \(x_j\) is not needed.
  This observation reveals a novel insight about learning: Pairwise summary on data is sufficient for learning the optimal hidden representations.

  Further, \cite{duan2021modularizing} demonstrated that the output module can be trained well with as few as a single randomly chosen fully-labeled example per class.
  In practice, this indicates that it suffices to have annotators mostly provide labels of the form \(\mathbbm{1}_{\{y_i = y_j\}}\), i.e., identify if each given pair is from the same class, which may be easier to annotate compared to full labels \(y_i, y_j\).
  Further, being able to learn effectively without knowing the specific values of \(y_i, y_j\) may be useful for user privacy protection when the label contains sensitive information.
  The theory behind learning with such pairwise summary on data and its applications has recently been developed in \cite{duan2021labels,shimada2021classification}.

  \item \textbf{Optimization:}
  Since Proxy Objective methods completely decouple the trainings of the modules, each training session will be an optimization problem with a smaller set of parameters and potentially more desirable properties.
  For example, suppose the output layer is trained by itself, the optimization can be convex depending on the choice of the loss function.
  Studying modular learning through the lens of optimization can be a worthwhile future direction.
  \item \textbf{Connections with contrastive learning:}
  The Proxy Objective instantiation proposed in \cite{duan2021modularizing} (two-module version) can be viewed as a supervised analog of contrastive learning \cite{saunshi2019theoretical}.
  Specifically, contrastive learning also trains the network as two modules and uses a contrastive loss to train the hidden layers.
  A typical contrastive loss encourages the hidden layers to map similar example pairs closer and dissimilar pairs farther.
  The key difference arises from the fact that the two methods work in different learning settings, one (the standard contrastive learning) in unsupervised learning and the other in supervised learning.
  In unsupervised contrastive learning, a pair is considered similar or dissimilar based on some prior knowledge such as the proximity of the individual examples.
  On the other hand, in \cite{duan2021modularizing}, a pair is considered similar or dissimilar based on if the examples are from the same class.
  Due to their close relationship, we expect results and observations from contrastive learning to benefit the research in Proxy Objective methods, and vice versa.

  \cite{lowe2019putting} directly applied self-supervised contrastive learning as a proxy objective for training the hidden layers.
  However, the optimality of such practice was not justified in the paper.
\end{itemize}

\subsection{Weakly Modular Learning Schemes}
\label{sec3:ne2e}

\subsubsection{Target Propagation}
\label{sec3:tp}

On a high level, Target Propagation methods train each module by having it regress to an assigned target.
This target is chosen such that perfect regression to it results in a decrease in the overall loss function value or at least a decrease in the loss function value of the immediate downstream module.
At each training step, the targets for the network modules are found sequentially (from the output module to the input module).
The output module uses the true labels as its target.

Specifically, Target Propagation assumes the usage of an iterative optimization algorithm, and, at each optimization step, generates a target \(t_1\)\footnote{This target depends on the label and network parameters \(\theta_1, \theta_2\), but is considered fixed once computed during the actual optimization. Therefore, this dependence is not explicitly written out.} for \(f_1(\cdot, \theta_1)\) to regress to, thereby removing the need for end-to-end backward pass (of gradients).
This target is essentially chosen as the inverse image of the label under \(f_2(\cdot, \theta_2)\).
And to approximate this inverse, an auxiliary model \(g_2(\cdot, \gamma_2)\), typically chosen to be a fully-connected neural network layer, needs to be learned alongside the main model \(f\), where \(\gamma_2\) represents the trainable parameters of \(g_2\).
Target Propagation methods can be understood as backpropagating the target through approximated inverses of layers.
Re-writing \(L\left(f, \theta_1, \theta_2, \left(x, y\right)\right)\) as \(H\left(f\left(x, \theta_1, \theta_2\right), y, \theta_1, \theta_2\right)\) for some \(H\), these methods can be abstracted as Algorithm~\ref{alg:tp}.
Note that this presentation assumes training batch size being set to \(1\).
An extension to mini-batch training with batch size greater than \(1\) is trivial.

\begin{algorithm}[t]
  \caption{An Optimization Step in Target Propagation}
  \label{alg:tp}
  \begin{algorithmic}[1]
    \Require An auxiliary inverse model \(g_2(\cdot, \gamma_2)\), a target generating function \(T\left(g_2\left(\cdot, \gamma_2\right), u, v, w\right)\) (\(u, v, w\) are placeholder variables), training data \(\left(x_{i}, y_{i}\right)\in S\), a reconstruction loss \(\ell^\text{rec}\left(g_2\left(\cdot, \gamma_2\right), f\left(\cdot, \theta_1, \theta_2\right), x_{i}\right)\), step size \(\eta > 0\)
    \Begin
    \State End-to-end forward pass 
    \begin{equation}
      a_1\left(\theta_1\right) \leftarrow f_1\left(x_{i}, \theta_1\right), a_2\left(\theta_1, \theta_2\right) \leftarrow f_2\left(a_1\left(\theta_1\right), \theta_2\right)
    \end{equation}
    \State Obtain target for output module 
    \begin{equation}
      t_2 \leftarrow a_2\left(\theta_1, \theta_2\right) - \eta\frac{\partial H(u, y_{i}, \theta_1, \theta_2)}{\partial u}\bigg\vert_{u = a_2\left(\theta_1, \theta_2\right)}  
    \end{equation}
    \State Obtain target for input module 
    \begin{equation}
      t_1 \leftarrow T\left(g_2\left(\cdot, \gamma_2\right), a_1\left(\theta_1\right), a_2\left(\theta_1, \theta_2\right), t_2\right)
    \end{equation}
    \State Update \(\gamma_2\) to minimize 
    \begin{equation}
      \ell^\text{rec}\left(g_2\left(\cdot, \gamma_2\right), f\left(\cdot, \theta_1, \theta_2\right), x_{i}\right)
    \end{equation}
    \State Update \(\theta_1\) to minimize 
    \begin{equation}
      \left\|a_1\left(\theta_1\right) - t_1\right\|_2^2
    \end{equation}
    \State Fix \(\theta_1\), update \(\theta_2\) to minimize 
    \begin{equation}
      \left\|a_2\left(\theta_1, \theta_2\right) - t_2\right\|_2^2
    \end{equation}
    \End
  \end{algorithmic}
\end{algorithm}

Target propagation does not allow fully modularized training since forward pass is always needed in each training step.

\paragraph{Instantiations}

Different instantiations of Target Propagation differ in their choices of the target generation function \(T\) and the reconstruction loss \(\ell^\text{rec}\).

\begin{itemize}
  \item Vanilla Target Propagation \cite{bengio2014auto,lee2015difference}:
  \begin{align}
    &T\left(g_2\left(\cdot, \gamma_2\right), a_1\left(\theta_1\right), a_2\left(\theta_1, \theta_2\right), t_2\right) := g_2\left(t_2, \gamma_2\right);\\ 
    &\ell^\text{rec}\left(g_2\left(\cdot, \gamma_2\right), f\left(\cdot, \theta_1, \theta_2\right), x_{i}\right) \\\nonumber
    & \quad := \left\|g_2\left(f_2\left(a_1\left(\theta_1\right) + \epsilon, \theta_2\right), \gamma_2\right) - \left(a_1\left(\theta_1\right) + \epsilon\right)\right\|_2^2, 
  \end{align}
  where \(\epsilon\) is some added Gaussian noise and \(a_1, a_2\) are stashed values from forward pass (see Algorithm \ref{alg:tp}).
  It is easy to see how this reconstruction loss encourages \(g_2\) to approximate an ``inverse'' of \(f_2\), and the added noise \(\epsilon\) enhances generalization.
  In the case where the network is trained as \(Q\) \((Q \geq 2)\) modules, the reconstruction loss for \(g_q\), \(q=2, \ldots, Q\) becomes
  \begin{align}
    &\ell^\text{rec}_q\left(g_q\left(\cdot\right), f\left(\cdot\right), x_{i}\right) \\\nonumber
    & \quad := \left\|g_q\left(f_q\left(a_{q-1} + \epsilon\right)\right) - \left(a_{q-1} + \epsilon\right)\right\|_2^2, 
  \end{align}
  where the trainable parameters are omitted for simplicity.
  \item Difference Target Propagation \cite{lee2015difference}:
  \begin{align}
    &T\left(g_2\left(\cdot, \gamma_2\right), a_1\left(\theta_1\right), a_2\left(\theta_1, \theta_2\right), t_2\right) \\\nonumber
    & \quad := g_2\left(t_2, \gamma_2\right) + \left[a_1\left(\theta_1\right) - g_2\left(a_2\left(\theta_1, \theta_2\right), \gamma_2\right)\right].
  \end{align}
  Difference Target Propagation uses the same reconstruction loss as Vanilla Target Propagation.
  The extra term \(\left[a_1\left(\theta_1\right) - g_2\left(a_2\left(\theta_1, \theta_2\right), \gamma_2\right)\right]\) is to correct any error \(g_2\) makes in estimating an ``inverse'' of \(f_2\).
  And it can be shown that this correction term enables a more robust optimality guarantee.
  \item Difference Target Propagation with Difference Reconstruction Loss \cite{meulemans2020theoretical}:
  This instantiation uses the same target generating function \(T\) as Difference Target Propagation but a different reconstruction loss dubbed the Difference Reconstruction Loss (DRL).
  Define difference corrected \(g_2\) as \(g'_2(\cdot, \gamma_2) = g_2\left(\cdot, \gamma_2\right) + \left[a_1\left(\theta_1\right) - g_2\left(a_2\left(\theta_1, \theta_2\right), \gamma_2\right)\right]\), the DRL is given as
  \begin{align}
    &\ell^\text{rec}\left(g_2\left(\cdot, \gamma_2\right), f\left(\cdot, \theta_1, \theta_2\right), x_{i}\right) \\\nonumber
    & \quad := \left\|g'_2\left(f_2\left(a_1\left(\theta_1\right) + \epsilon_1, \theta_2\right), \gamma_2\right) - \left(a_1\left(\theta_1\right) + \epsilon_1\right)\right\|_2^2\\\nonumber
    & \quad\quad + \lambda\left\|g'_2\left(a_2\left(\theta_1, \theta_2\right) + \epsilon_2, \gamma_2\right) - a_1\left(\theta_1\right)\right\|_2^2
  \end{align}
  where \(\epsilon_1, \epsilon_2\) are some added Gaussian noise and \(\lambda\) a regularization parameter.
  In the case where the network is trained as \(Q\) modules, the reconstruction loss for \(g_q\) becomes
  \begin{align}
    &\ell^\text{rec}_q\left(g_q, \ldots, g_Q, f_{q\rightarrow Q}, x_{i}\right) \\\nonumber
    & \quad := \left\|g'_q\circ\cdots\circ g'_Q\left(f_{q\rightarrow Q}\left(a_{q-1} + \epsilon_1\right)\right) - \left(a_{q-1} + \epsilon_1\right)\right\|_2^2\\\nonumber
    & \quad\quad + \lambda\left\|g'_q\circ\cdots\circ g'_Q\left(a_Q + \epsilon_2\right) - a_{q-1}\right\|_2^2,
  \end{align}
  where the trainable parameters are omitted for simplicity, \(g'_q\circ\cdots\circ g'_Q\) (by abusing notation) denotes the operation of recursively computing inverse until \(g'_q\), and \(f_{q\rightarrow Q}\) denotes the composition of \(f_q, \ldots, f_Q\).
  \item Direct Difference Target Propagation \cite{meulemans2020theoretical}:
  Modify the Difference Reconstruction Loss by modeling \(g'_q\circ\cdots\circ g'_Q\) with a direct learnable connection from the output module into the activation space of module \(q-1\).
\end{itemize}

\paragraph{Optimality Guarantees}

If \(g_2=f_2^{-1}\) and \(f_2\) is a linear layer on top of some elementwise nonlinearity, then the local gradient produced by Vanilla Target Propagation can be shown to align well with (within a \(90^\circ\) proximity of) the true gradient of the overall loss function in the input module under some additional mild assumptions (Theorem 1 in \cite{lee2015difference}).
A similar result can be found in \cite{meulemans2020theoretical} (Theorem 6).

Under some mild assumptions, the target produced by Difference Target Propagation can be shown to cause a decrease in the loss of the immediate downstream module if this downstream module is already close to its own target (Theorem 2 in \cite{lee2015difference}).

In \cite{meulemans2020theoretical}, it was shown that, because \(f_2\) is typically not invertible, Difference Target Propagation does not propagate the so-called ``Gauss-Newton target'' as \(t_1\), i.e., target that represents an update from an approximate Gauss-Newton optimization step.
And minimizing the proposed DRL encourages the propagation of such a target.
The benefit of training the input module to regress to a Gauss-Newton target is that, at least in certain settings, the resulting gradients in the input module can be shown to align well with gradients computed from the overall loss (Theorem 6 in \cite{meulemans2020theoretical}), thus leading to effective training.

Almost all existing theoretical optimality results require that the network consists of purely linear layers (possibly linked by elementwise nonlinearities).

\paragraph{Advantages}

As a weakly modular method, Target Propagation does not require end-to-end backward pass and updates each module individually after each forward pass is done.
When computing gradients for the entire model using the chain rule becomes expensive, Target Propagation may therefore help save computations since it only needs module-wise gradients.
This also enables training models that have non-differentiable operations, as demonstrated in \cite{lee2015difference}.
These advantages are of course shared by the other two families of methods.

Note that if and how much computation can be saved highly depends on the actual use case.
To be specific, while Target Propagation does not need full backward pass, it does require nontrivial computations for calculating the hidden targets that involve training and evaluating the auxiliary models.
Depending on how complex these models are, this extra workload may overweigh the saving of eliminating full backward pass.
Further, some Target Propagation variants require stashing forward pass results for evaluating hidden targets, meaning that they may not have any advantage over E2EBP in terms of memory footprint either.

\paragraph{Current Limitations and Future Work}

\begin{itemize}
  \item The auxiliary models require extra human (architecture selection, hyperparameter tuning, etc.) and machine resources. 
  \item Target Propagation methods have not been shown to yield strong performance on more challenging datasets such as CIFAR-10 and ImageNet or on more competitive networks \cite{meulemans2020theoretical,bartunov2018assessing}.
  \item Similar to Proxy Objective methods, optimality results for more general settings, in particular, broader network architecture families, are lacking.
\end{itemize}






\subsubsection{Synthetic Gradients}
\label{sec3:sg}

Synthetic Gradients methods approximate local gradients and use those in place of real gradients produced by end-to-end backward pass for training.
Specifically, Synthetic Gradients methods assume that the network weights are updated using a gradient-based optimization algorithm such as stochastic gradient descent.
Then these methods approximate local gradients with auxiliary models.
These auxiliary models are typically implemented with fully-connected networks, and are trained to regress to a module's gradients (gradients of the overall objective function with respect to the module's activations) when given its activations.
By leveraging these local gradient models, Synthetic Gradients methods reduce the frequency in which end-to-end backward passes are needed by using the synthesized gradients in place of real gradients. 
End-to-end backward passes are only performed occasionally to collect real gradients for training the local gradient models.
Re-writing \(L\left(f, \theta_1, \theta_2, \left(x, y\right)\right)\) as \(H\left(f\left(x, \theta_1, \theta_2\right), y, \theta_1, \theta_2\right)\) for some \(H\), these methods can be abstracted as Algorithm~\ref{alg:synthetic}.

It is possible to reduce the frequency that the forward pass is needed as well by approximating the forward pass signals with auxiliary synthetic input models that predict inputs to modules given data.

Synthetic Gradients methods do not allow truly weakly modular training since occasional end-to-end backward (or forward) passes are needed to learn the synthetic gradient (or input) models.
They can only be used to accelerate end-to-end training and enable parallelized optimizations.

\begin{algorithm}[t]
  \caption{An Optimization Step in Synthetic Gradients}
  \label{alg:synthetic}
  \begin{algorithmic}[1]
    \Require An auxiliary synthetic gradient model \(s_2\left(\cdot, \psi_2\right)\), training data \(\left(x_{i}, y_{i}\right)\), step size \(\eta > 0\)
    \Begin
    \State End-to-end forward pass 
    \begin{equation}
      a_1\left(\theta_1\right) \leftarrow f_1\left(x_{i}, \theta_1\right), a_2\left(\theta_1, \theta_2\right) \leftarrow f_2\left(a_1\left(\theta_1\right), \theta_2\right)
    \end{equation}
    \State Update output module 
    \begin{equation}
      \theta_2 \leftarrow \theta_2 - \eta\frac{\partial H\left(f\left(x_i, \theta_1, u\right), y_i, \theta_1, u\right)}{\partial u}\bigg\vert_{u = \theta_2}
    \end{equation}
    \State Obtain synthetic gradients for input module 
    \begin{equation}
      \hat{\delta}_1\left(\theta_1, \psi_2\right) \leftarrow s_2\left(a_1\left(\theta_1\right), \psi_2\right)
    \end{equation}
    \State Update input module 
    \begin{equation}
      \theta_1 \leftarrow \theta_1 - \eta \hat{\delta}_1\left(\theta_1, \psi_2\right)\frac{\partial a_1\left(u\right)}{\partial u}\bigg\vert_{u=\theta_1}
    \end{equation}
    \If{update synthetic gradient model}
      \State End-to-end backward pass, obtain true gradients 
      \begin{equation}
        \delta_1 \leftarrow \frac{\partial H\left(f_2\left(u, \theta_2\right), y_{i}, \theta_1, \theta_2\right)}{\partial u}\bigg\vert_{u=a_1\left(\theta_1\right)}
      \end{equation}
    \State Update \(\psi_2\) to minimize 
    \begin{equation}
      \left\|\delta_1 - \hat{\delta}_1\left(\theta_1, \psi_2\right)\right\|_2^2
    \end{equation}
    \EndIf
    \End
  \end{algorithmic}
\end{algorithm}

\paragraph{Instantiations}
\begin{itemize}
  \item \cite{jaderberg2017decoupled,czarnecki2017understanding}: Proposed the original instantiation, which is fully described above. 
  \item To remove the need for occasional end-to-end backward pass, \cite{lansdell2020learning} proposed a method to obtain target signals for training the synthetic gradient models using only local information.
  However, this method necessitates the use of stochastic networks. 
  And the performance reported in the paper is underwhelming compared to either \cite{jaderberg2017decoupled} or E2EBP. 
  Therefore, we do not consider it as a training method that works in the general set-up as the rest of the methods in this paper do, and it is included here only for completeness.
\end{itemize}

\paragraph{Optimality Guarantees}

Optimality can be trivially guaranteed assuming that the synthetic gradient models perfectly perform their regression task, i.e., that they perfectly approximate local gradients using module activations.
However, this assumption is almost never satisfied in practice.

\paragraph{Advantages}

\begin{itemize}
  \item While Proxy Objective and Target Propagation methods all help reduce computational load and memory usage, Synthetic Gradients methods can enable parallelized training of network modules. 
  This can further reduce training cost since update on a certain module does not need to wait for those of other modules except when the synthetic input or synthetic gradient models are being updated.

  This advantage is shared by a variant of Proxy Objective \cite{belilovsky2020decoupled}.
  And it was shown in \cite{belilovsky2020decoupled} that this Proxy Objective variant is much more performant than Synthetic Gradients.

  \item Synthetic Gradients can be used to approximate true backpropagation through time (unrolled for an unlimited number of steps) for learning recurrent networks.
  It was shown in \cite{jaderberg2017decoupled} that this allows for much more effective training for learning long-term dependency compared to the usual truncated backpropagation through time.
\end{itemize}

\paragraph{Current Limitations and Future Work}

\begin{itemize}
  \item Similar to Target Propagation, the auxiliary models require extra human and machine resources.
  And like Target Propagation, there is no empirical evidence that Synthetic Gradients can scale to challenging benchmark datasets or competitive models.
  \item Synthetic Gradients methods do not enable truly weakly modular training in general.
\end{itemize}

\section{Other Non-E2E Training Approaches}
\label{other_methods}

\subsection{\rch{Methods Motivated Purely by Biological Plausibility}}
Arguably the most notable set of works left out in this survey are those studying alternatives to E2EBP purely from the perspective of biological plausibility (for a few examples, see \cite{balduzzi2014kickback,xiao2018biologically,nokland2016direct,lillicrap2016random,liao2016important}).
These training methods were generally purely motivated by our understandings on how human brain works, and were therefore claimed to be more ``biological plausible'' than E2EBP.
However, biological plausibility in itself does not lead to provable optimality.
Despite the fact that these methods are of great value from a biology standpoint, they have been significantly outperformed by E2EBP on meaningful benchmark datasets \cite{bartunov2018assessing}.
\rch{Below, arguably the most popular family of biologically plausible alternatives to E2EBP -- Feedback Alignment -- is discussed.}

\rch{One major argument criticizing E2EBP's lack of biological plausibility states that E2EBP requires each neuron to have precise knowledge of all of the downstream neurons (end-to-end backward pass), whereas human brain is not believed to demonstrate such precise pattern of reciprocal connectivity \cite{lillicrap2016random}. 
This issue is known as the ``weight transport'' problem \cite{lillicrap2016random}.
}
\rch{To solve the weight transport problem, \cite{lillicrap2016random} proposed to use fixed, random weights in place of actual network weights during backward pass, breaking the symmetry between weights used during forward and backward passes and thus solving the problem.
This eliminates the need for a true end-to-end backward pass.
And this family method is called Feedback Alignment.
\cite{liao2016important} proposed to use fixed, random weights that share signs with the actual network weights.
\cite{nokland2016direct} proposed two more revisions of the original Feedback Alignment instantiation.
During the backward pass, instead of using the backpropagated supervision (with random weights instead of real weights) to provide gradient like the original instantiation does, these alternative versions directly use error at the output (with potential modulation by random matrices to make the dimensionality match for each layer).
}

\rch{Suppose the model is written as \(f(x, W_1, W_2, W_3) = \sigma_3\left(W_3\sigma_2\left(W_2\sigma_1\left(W_1 x\right)\right)\right) = f_3(f_2(f_1(x, W_1), W_2), W_3)\), where \(W_1, W_2, W_3\) are trainable weight matrices and \(\sigma_1, \sigma_2, \sigma_3\) are activation functions.
During any step in gradient descent, suppose the forward pass has been done and let \(a_3 = f(x, W_1, W_2, W_3), a_2 = f_2(f_1(x, W_1), W_2), a_1 = f_1(x, W_1), b_2 = W_2 a_1\), where \(W_1, W_2\) are the current network weights, E2EBP computes gradient in \(f_1\) as}
\begin{equation}
  \rch{\frac{\partial{L}}{\partial{a_1}} = \frac{\partial{L}}{\partial{a_3}}\frac{\partial{a_3}}{\partial{a_2}}\frac{\partial\sigma_2}{\partial b_2}W_2.} 
\end{equation}
\rch{The original instantiation of Feedback Alignment in \cite{nokland2016direct} computes this gradient with \(W_2\) substituted by some fixed, random matrix \(B_2\).
The variant proposed in \cite{liao2016important} substitutes \(W_2\) with fixed, random matrix \(B_2\) with the only constraint being that each element of \(B_2\) shares the same sign with the corresponding element in \(W_2\).
\cite{nokland2016direct} proposed to compute this gradient as
}
\begin{equation}
  \rch{\frac{\partial{L}}{\partial{a_1}} = \frac{\partial{L}}{\partial{a_3}}\frac{\partial\sigma_2}{\partial b_2}C_2,}
\end{equation}
\rch{where \(C_2\) is a fixed, random matrix with appropriate dimensionality.
The error at the output (\(\partial L / \partial a_3\)), after modulation by some random matrix \(C_2\), is used for training in place of backpropagated supervision.}

\subsection{\rch{Auxiliary Variables}}
Another important family of E2EBP-free training methods is the Auxiliary Variables methods \cite{carreira2014distributed,taylor2016training,marra2020local,raghavan2020distributed,gu2020fenchel,zhang2017convergent,askari2018lifted,lau2018proximal,li2019lifted,zhang2016efficient,zeng2019global}. 
These methods introduce auxiliary trainable variables that approximate the hidden activations in order to achieve parallelized training. 
Despite the strong theoretical guarantees, the introduced auxiliary variables may pose scalability issues and, more importantly, these methods require special, often tailor-made alternating solvers. 
And none of the existing work in this area has scaled these methods to go beyond toy models and toy data. 
\rch{The basic idea and connections with the reviewed methods will be discussed below.}

\rch{On a high level, Auxiliary Variables methods are built upon the idea of variable splitting, that is, transforming a complicated problem in which variables are coupled highly nonlinearly into a simpler one where the coupling between variables becomes more tractable by introducing additional variables \cite{zeng2019global}. 
Auxiliary Variables methods assume a representation of the model as follows: \(f(x, W_1, W_2) = \sigma_2\left(W_2\sigma_1\left(W_1 x\right)\right)\), where \(W_1, W_2\) are trainable weight matrices and \(\sigma_1, \sigma_2\) are activation functions. 
Suppose the objective function \(L\left(f, W_1, W_2, S\right)\) can be written as \(\sum_{i=1}^nL\left(f, W_1, W_2, \left(x_i, y_i\right)\right)\). 
Re-writing \(L\left(f, W_1, W_2, \left(x, y\right)\right)\) as \(H\left(f\left(x, W_1, W_2\right), y, W_1, W_2\right)\) for some \(H\), one popular family of Auxiliary Variables methods amounts to introducing new variables \(V_1, V_2\), and reformulating the training objective as follows:}
\begin{align}
  &\rch{\min_{W_1, W_2}\sum_{i=1}^n H\left(f\left(x_i, W_1, W_2\right), y_i, W_1, W_2\right)} \\
  &\rch{\rightarrow \min_{W_1, W_2, V_1, V_2}\sum_{i=1}^n H'\left(V_2\left(V_1\left(x_i\right)\right), y_i, W_1, W_2, V_1, V_2)\right)} \nonumber\\
  & \rch{\text{subject to }} \nonumber\\
  &\,\,\rch{V_1\left(x_i\right) = \sigma_1\left(W_1 x_i\right), V_2\left(V_1\left(x_i\right)\right) = \sigma_2\left(W_2 V_1\left(x_i\right)\right), \forall i.}
\end{align}
\rch{One example of \(H\left(f\left(x_i, W_1, W_2\right), y_i, W_1, W_2\right)\) could be the hinge loss on \(\left(f\left(x_i, W_1, W_2\right), y_i\right)\) plus some regularization terms on the weights \(W_1, W_2\), in which case \(H'\left(V_2\left(V_1\left(x_i\right)\right), y_i, W_1, W_2, V_1, V_2)\right)\) is the hinge loss on \(\left(V_2\left(V_1\left(x_i\right)\right), y_i\right)\) plus regularization terms on \(W_1, W_2, V_1, V_2\).
This constrained optimization problem is typically turned into an unconstrained one by adding a regularization term forcing \(V_2\left(\cdot\right)\) to regress to \(\sigma_2\left(W_2\, \cdot\right)\) and \(V_1\left(\cdot\right)\) to \(\sigma_1\left(W_1\, \cdot\right)\).
The variables \((W_1, W_2) \text{ and } (V_1, V_2)\) are solved alternatively.
Popular solvers include block coordinate descent \cite{carreira2014distributed,zeng2019global,zhang2017convergent,lau2018proximal,askari2018lifted} and alternating direction method of multipliers \cite{taylor2016training,zhang2016efficient}. 
Strong convergence proofs on these solvers are provided in, e.g., \cite{zeng2019global}.}

\rch{Another popular family of Auxiliary Variables methods introduces another set of auxiliary variables: \(V_i\left(\cdot\right) = \sigma_i\left(U_i(\cdot)\right), U_i(\cdot) = W_i\,\cdot, i = 1, 2\).
This instantiation is even more tractable than the previous one, often enjoying closed-form solutions.
The drawback is the additional computation and memory introduced by the third set of auxiliary variables \(U_i, i = 1, 2\).
}

\rch{Auxiliary Variables methods are conceptually similar to Target Propagation methods in the sense that they eliminate end-to-end backward pass by introducing new, local-in-each-layer targets for the hidden layers to regress to.
Training then alternates between solving these targets and optimizing the main network weights using local supervision.
Target Propgagation leverages (approximated) inverse images of the label through the layers as the hidden targets.
In comparison, the hidden targets (auxiliary variables) used in Auxiliary Variables methods can be interpreted as approximations of the forward images of the input through the layers.}

\section{Conclusion}
\rch{This survey} reviewed modular and weakly modular training methods for deep architectures as alternatives to the traditional end-to-end backpropagation.
These alternatives can match or even surpass the performance of end-to-end backpropagation on challenging datasets like ImageNet.
In addition, they are natural solutions to some of the practical limitations of end-to-end backpropagation and they reveal novel insights about learning.
As the interest in modular and weakly modular training schemes continues to grow, we hope that this short survey can serve as a summary of the progress made in this field and inspire future research.

\section*{Acknowledgment}
This work was supported by the Defense Advanced Research Projects Agency (FA9453-18-1-0039) and the Office of Naval Research (N00014-18-1-2306).

\bibliographystyle{IEEEtran}
\bibliography{main}

\end{document}